# Trained Model Fusion for Object Detection using Gating Network

Tetsuo Inoshita, Yuichi Nakatani, Katsuhiko Takahashi,
Asuka Ishii, Gaku Nakano

NEC Corporation
{tetsuo.inoshita, y.nakatani, katsuhiko.takahashi,
as-ishii, g-nakano}@nec.com

**Abstract.** The major approaches of transfer learning in computer vision have tried to adapt the source domain to the target domain one-to-one. However, this scenario is difficult to apply to real applications such as video surveillance systems. As those systems have many cameras installed at each location regarded as source domains, it is difficult to identify the proper source domain. In this paper, we introduce a new transfer learning scenario that has various source domains and one target domain, assuming video surveillance system integration. Also, we propose a novel method for automatically producing a high accuracy model by fusing models trained at various source domains. In particular, we show how to apply a gating network to fuse source domains for object detection tasks, which is a new approach. We demonstrate the effectiveness of our method through experiments on traffic surveillance datasets.

## 1 Introduction

Along with the development of deep learning, the number of video surveillance systems with installations of hundreds of surveillance cameras is increasing. For improving recognition accuracy at new camera installation locations (target domain), for deep learning, it is helpful to use data at locations where cameras are already installed (source domains). However, as the data distribution is different at each camera installation location, finding the most suitable data from source domains is difficult.

Transfer learning aims to close the distribution gap between such different domains. These underlying technologies are to shift the data distribution of different domains so that models trained in the source domain can be used in the target domain. Fine-tuning is a simple and powerful way to use a pre-trained model [1]. The method is to fine-tune an existing network that was trained on a large dataset, such as the ImageNet [2], by continuing to train it (i.e., run back-propagation) on the smaller dataset we have.

In many typical approaches to transfer learning, a major assumption is that the source and target domain are transferred one-to-one by using an image classification task. Recently, as a more practical scenario, Z. Cao [3, 4] proposed partial adversarial domain adaptation, which assumes that the source label space is a super space of the target label space. It can transfer knowledge from a big domain of many labels (e.g., ImageNet [2], Google Open Image [5]) to a small domain of few labels.

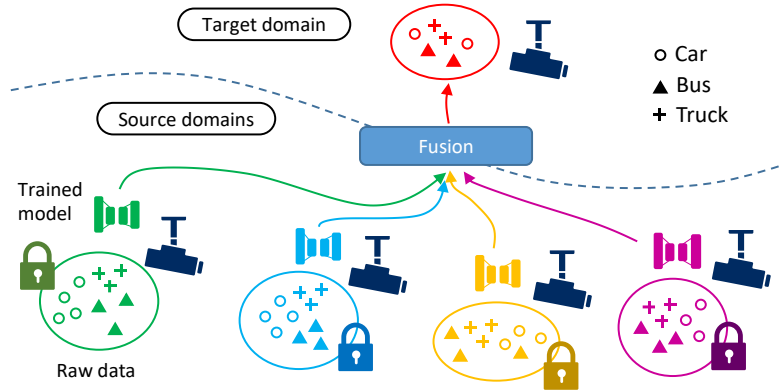

**Fig. 1.** Our transfer learning scenario supposes that the knowledge is transferred from various source domains to a target domain. Assuming that there is a video surveillance system, there are many source domains that consist of images from cameras installed at each location. However, due to privacy issues, we cannot use those images. Therefore, we handle the trained models that are generated by images and consist of parameters with a low privacy risk. Our proposed method produces a suitable model for a target domain by fusing trained models of these source domains.

However, in real applications, coping with the previous transfer learning scenarios is difficult. In the case of video surveillance systems, there are many live cameras that can be considered as multiple source domains. In this situation, transfer learning approaches require selecting suitable source domains for a target domain, but it is not easy. As each source domain consists of data generated by various camera locations, random selection from source domains may not be valid for a target domain and manual selection takes much time. Therefore, in this work, we focus on producing a good model for a target domain by using various source domains automatically, assuming the scenario of video surveillance system integration. Also, existing transfer learning approaches often deal with raw data, such as images or videos, as the source domain. However, storing these raw data is a high risk in terms of privacy. Instead, we utilize trained models that contain network parameters, not personal information. This trained model characteristic is effective in social implementation.

The challenge of this new scenario is to create a model suitable for a target domain with trained models that have various data distribution features that are generated in many environments (Fig. 1).

In this paper, we propose a fusing method for producing a high accuracy model for a target domain by using various trained models with different source domains. To accomplish this goal, we use the gating network, which was originally proposed using a mixture of experts [6, 7]. Basically, the previously proposed gating network aims to solve classification tasks. However, a method that applies to object detection tasks that estimate the location information of objects in an image has not been reported.

To summarize, our main contributions are as follows: we (1) introduce a new transfer learning scenario that assumes that there is video surveillance system integration and (2) design a method to optimally fuse various trained models that are trained in different source domains using the gating network, especially for object detection tasks. Our experimental results show that our proposed method can achieve better detection accuracy than previous methods.

## 2 Related work

### 2.1 Ensemble of Experts

Generally, an ensemble of multiple experts (e.g., pre-trained model, neural network, classifier, and detector) is a well-known technique to improve recognition accuracy [9, 10, and 11]. Experts are used in many ways depending on the purpose. In this section, we define expert as three types (Fig. 2) as detailed below.

(1) detector-specific experts: These experts consist of various types of object detectors. In the studies of S. Bae et al. and G. Zhou et al [12, 16], they provide a method for building an ensemble detection to combine different types of detectors (e.g., Faster R-CNN [13], SSD [14], DSSD [15], and YOLOv2 [17]). They demonstrate the effectiveness of their approach.

(2) class-specific experts: The target class of recognition is different for each expert. In [31], they design a multi-teacher knowledge distillation framework for the image classification task. Their purpose is to merge the knowledge from a lot of teachers (i.e., experts on cars, buses, trucks, birds, and flowers) into a single student model.

(3) bias-specific experts: We introduce a new definition of expert. The bias means the difference of data acquisition environment depending on the location, e.g. lighting, camera angles, and background. In video surveillance systems, for improving the accuracy, an expert is often trained by using the bias data of the installed location. Therefore, an expert becomes an expert specifically for the location. In this paper, our scenario defines the expert as the bias-specific expert.

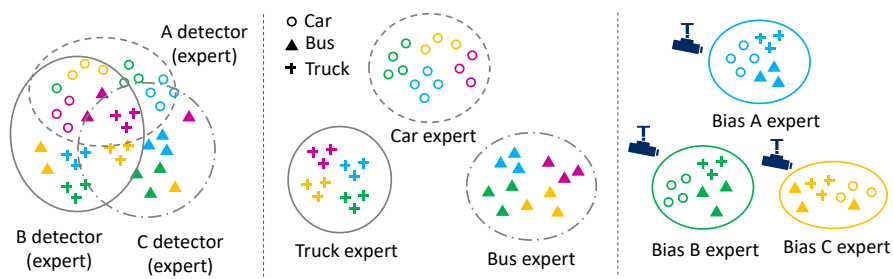

**Fig. 2.** (Left) detector-specific expert, (Center) class-specific expert, (Right) bias-specific expert. In this work, we deal with the bias-specific experts that are trained in different places. These experts have the same classes among experts, but differ from captured situations.

## 2.2 Gating Network

The gating network performs the important function of adjusting the fusion weights among the experts. The mixture of experts [6, 7] is one of the most popular methods that controls the activation of multiple experts using a gating network. This is achieved by training a gating network that maps each input to a distribution over the experts. These gating network architectures have been applied to various fields such as language models and machine translation [18, 19, and 20].

In the field of computer vision, many applications using a gating network have been proposed [21]. J. Kim [22] introduced a deep fusion network that controls the amount of information coming from each modality through a gating network. S. Dodge et al. [23] proposed a visual saliency model that is formulated as a mixture of experts. The saliency map is computed as a weighted mixture of experts with weights determined by a gating network.

In our scenario, video surveillance system integration, due to the varieties of source domain, it is not easy to decide a source domain that is suitable for the target domain manually. Therefore, we design a method to optimally fuse various trained models that are trained in different source domains using a gating network automatically, especially for object detection tasks.

## 2.3 Object Detection

Current state-of-the-art object detectors have been classified into two types: (1) two-stage detectors (e.g., R-CNN [24], Faster R-CNN [13]) first generate a sparse set of candidate object region proposals and then refine the accurate object regions and the corresponding class labels by using CNN networks and (2) one-stage detectors (e.g., SSD [14], YOLOv3 [25]) are applied over a regular, dense sampling of object location that is called *anchor* [13]. The strength of one-stage detectors is high computational efficiency, which is desired for real time applications. But the detection accuracies of one-stage detectors are usually worse than those of two-stage detectors.

Some recent one-stage approaches have tackled these problems. To be robust to scale changes, the feature pyramid network (FPN) [26] was proposed. It is a simple framework that has a pyramidal hierarchy of deep convolutional networks. Class imbalance is also a well-known big problem of one-stage detectors. To address this problem, Lin et al. [27] proposed focal loss as a loss function that focuses on training on a sparse set of hard samples and prevents a vast number of easy negatives.

In this work, we use RetinaNet [27] as an object detector that has the architecture of the FPN and the focal loss. It consists of a base network and two task-specific prediction subnets. The role of the base network including the FPN is to extract features. The first subnet, the classification subnet, is used to classify objects and predict the probability of each anchor. The second subnet, the box regression subnet, is used to regress the bounding box position and the offset from each anchor box to the matched ground-truth object.

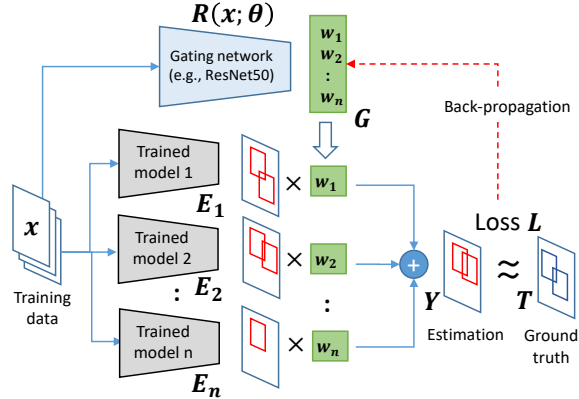

**Fig. 3.** The training framework of a gating network for object detection tasks. It has a function to fuse multiple trained models according to the weight. In the training phase, we calculate the weighted average of the output value of each trained model using the weight decided by the gating network. Then, we calculate the value of the loss function and re-train the gating network by simple back-propagation.

## 3   Proposed Method

### 3.1   Overall structure

In this work, we treat the expert as a trained model that is learned by using data collected at each source domain. Our overall training framework is illustrated in Fig. 3. We aim to produce a high accuracy model for object detection tasks by fusing multiple trained models that had already been generated in various source domains. We use the gating network as the fusion function. The following two paragraphs summarize the training and inference phase.

 **Training phase:** A gating network is trained to decide the suitable weight for each trained model. First, we build a gating network that has the same number of outputs ($w_n$) as the number of experts. Second, to estimate object location in an image with trained models, we calculate the weighted average by using the output value of each trained model and the weight decided by the gating network. Finally, we calculate the value of the loss function and give feedback to the gating network by simple back-propagation.

**Inference phase:** Object locations are predicted as bounding boxes in an image. First, an image is input to a gating network that has been already been trained, and then the fusion weights are acquired. Also, an image is input to each trained model, and then we stack the outputs that estimate the bounding boxes of objects. Second, we multiply the outputs by the weights, and then we apply non-maximum suppression (NMS) [28] for the outputs to remove overlapping boxes, which helps to reduce the number of unmerged false positives. After the NMS, we use the remaining boxes as final detections of the object.

### 3.2 Gating Network for object detection

Our gating network predicts, given an input image $x$, the value of the weight of each trained model $E(x)$. A simple gating function $G(x)$ is defined as

$$G(x) = softmax(R(x;\theta)) \quad (1)$$

where $R(x;\theta)$ is the deep neural network (e.g., VGG, ResNet-50) that replaced the original output layer with our defined layer that has the same number of outputs of trained models for a given input $x$ and $\theta$, which is a trainable parameter of a deep neural network. Let us denote the output of the gating network and the output of the i-th trained model by using $G(x)$ and $E_i(x)$. The output $Y$ through the gating network can be written as follows:

$$Y = \sum_{i=1}^{n} G(x)_i E_i(x) \quad (2)$$

As described in section 2.3, RetinaNet consists of a box regression subnet that has the locations of the objects in the image and a classification subnet that has the classification score set for each object location. Both subnets weighted by the gating network can be denoted as

$$Y_{reg} = \sum_{i}^{n} G(x)_i\, E_i(x)_{reg} \quad Y_{cls} = \sum_{i}^{n} G(x)_i\, E_i(x)_{cls}$$

### 3.3 Training Gating Network

In the training phase on the gating network, our method minimizes the joint loss function that consists of regression loss and classification loss

$$l_{reg} = l1\_smooth(Y_{reg}, T_{reg}) \quad (1)$$

$$l_{cls} = focal\ weight \cdot BCE(Y_{cls}, T_{cls}) \quad (2)$$

$$L = l_{reg} + l_{cls} \quad (3)$$

where the regression loss is the smooth L1 loss (l1_smooth) [29] between the prediction box $Y$ and the ground-truth box $T$.

**Top-k model selection:** If we have many trained models at the inference phase, we cannot apply these models to a real system due to a lack of computer resources. Therefore, we have to consider limiting the number of models used. We construct the gating network using the top k trained models instead of all of the trained models. Beforehand, we calculate the weight of the gating network using training data. Then, we select the top k trained models that have large weights. Finally, we re-train a gating network with selected models.

## 4 Experiments

Assuming traffic surveillance system integration, we evaluate the effectiveness of the proposed method by the detection accuracy of the vehicle. We consider the locations of the many surveillance cameras installed as source domains. Also, we consider the newly installed camera as the target domain.

### 4.1 Setup

**Dataset:** We use a part of the UA-DETRAC [30] dataset that contains 100 video sequences captured from real-world traffic scenes at different locations. In this work, we select 4 location videos as target domains (T1, T2, T3, and T4) for inference (Fig. 4). In this figure, for training, we are given a few target labeled data (T1', T2', T3', and T4'). As source domains, we use 30 location videos (S1 to S30) (Fig. 5). Also, we set the detection target to the *car* class. The dataset pattern we used in this experiment is shown in Table 1, and more details are given in Appendix A.

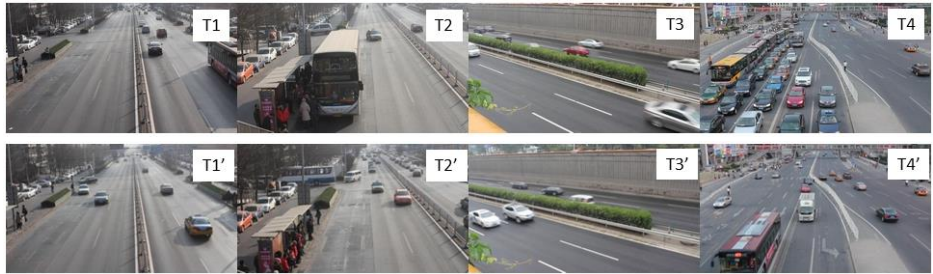

**Fig. 4.** Upper row: target domain (inference dataset), T1, T2, T3, and T4. Lower row: training dataset, T1', T2', T3', T4' for fine-tuning and for training the gating network. When we fine-tune the model, we usually use data that is similar to the target domain as much as possible. Therefore, in these experiments, we use data obtained from different camera angles that is similar but not identical.

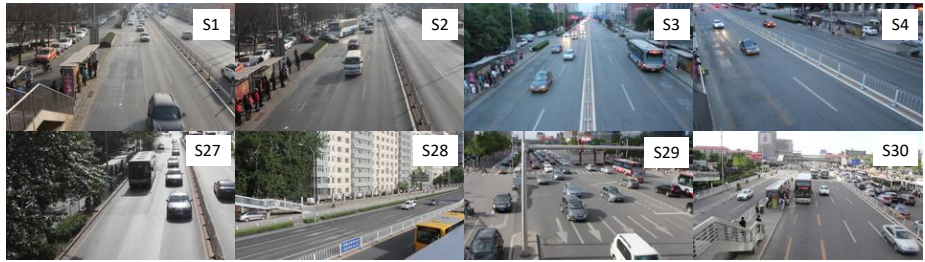

**Fig. 5.** A part of source domain dataset. Upper row: S1, S2, S3, and S4. Lower row: S27, S28, S29, and S30.

**Table 1.** Experimental datasets pattern

|  | Trained model | Fine-tuning | Average | Gating network |
|---|---|---|---|---|
| Training | S1, S2, …, S30 (source domains) | T1', T2', T3', T4' | | |
| Inference | T1, T2, T3, and T4 (target domains) | | | |

**Trained models:** In total, we produce 30 trained models with RetinaNet architecture using a source domain dataset. For example, the trained model S1 is trained by using the S1 dataset in Appendix A. Most hyperparameters are set to be the same as those of the original RetinaNet.

**Baseline:** <u>Fine-tuning:</u> Fine-tuning is one of the effective methods of transfer learning. Generally, fine-tuning needs a lot of labeled data for training. However, it is not desirable due to the huge cost of data annotation. In view of the situation, we use few training data for fine-tuning. The number of each training data (T1', T2', T3', and T4') is 44, 80, 115, and 103, respectively. We fine-tune the same object detector, RetinaNet, as for the trained models above.

**Baseline2:** <u>Average:</u> We simply calculate an average of the outputs from 30 trained models (S1 to S30). We then apply NMS to these outputs and acquire the estimated object locations. This experimental condition can be considered as the uniform weight of the gating network. Therefore, when we evaluate this average method, we use a gating network in which all the weights are set to 1.

**Proposed method 1:** <u>Gating Network (all):</u> The gating network architecture consists of ResNet-50 pre-trained on ImageNet as an initial value and an add output layer with the same number of outputs as the source domains. By using this architecture, we train the trainable weight of ResNet-50 with training data for the gating network. Under this condition, we use all the trained models as the elements of the gating network.

**Proposed method 2:** <u>Gating Network (top 5):</u> The network architecture is the same as for the above condition, but the number of trained models used for the gating network is different. We use the top 5 trained models according to *top-k model selection* (section 3.3). For example, the T1' dataset is input into the gating network that consists of all the trained models (S1 to S30). Then, we choose the top 5 models with the highest weight value. Using these 5 models, we re-train the gating network with the T1' dataset.

### 4.2 Results and Discussion

First, we evaluate the individual performance of the trained models considered as source domains. In Fig. 6, the accuracies for the T1, T2, T3, and T4 datasets are provided; more details are given in Appendix B. This shows that the trained models have their own strengths and weaknesses depending on the inference dataset. Thus, it is not easy to find one trained model (source domain) suitable for inference data.

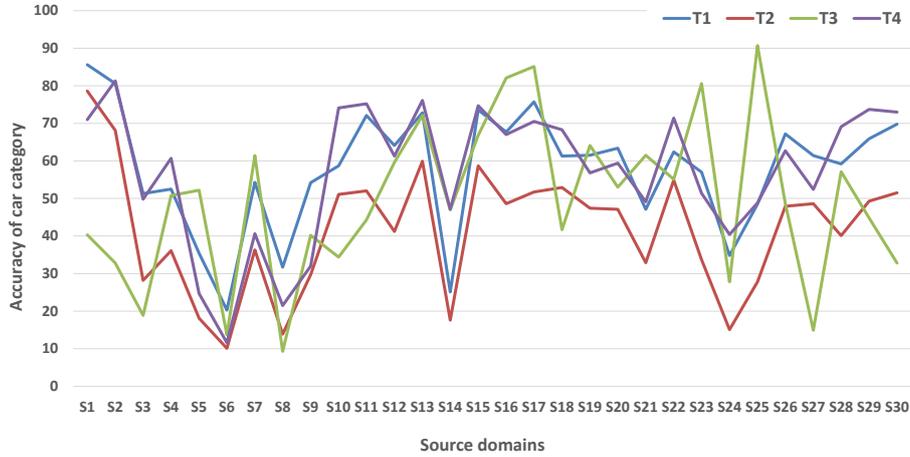

**Fig. 6.** Individual performance of 30 trained models. The accuracy of each source domain is different depending on the inference datasets (T1, T2, T3, and T4). Therefore, it is difficult to select one suitable trained model.

Next, we build a gating network using these trained models and compare the results with the results of the baselines with mean average precision (mAP) in Table 2. Judging the results of this experiment comprehensively, our proposed method (i.e., the gating network method) exceeds the baselines on several datasets. The following paragraphs discuss each individual method.

**Maximum accuracy of trained model:** Trained models have the potential to achieve higher accuracy than baselines. But, as mentioned before at the beginning of this section, there is a large difference in accuracy between the trained models. It is difficult to choose the best trained model for the inference dataset. Therefore, selecting one trained model is not a realistic method for real applications.

**Baseline fine-tuning:** For inference data, the overall accuracy of the fine-tuning method is lower than that of other methods. However, the accuracy for all training datasets is 100% (not listed in Table 2). Therefore, the cause of this decrease in accuracy for inference data is considered to be over-fitting. In this scenario, the amount of training data for fine-tuning is less than usual. If a large amount of training data can be prepared, this method may be good as usual.

**Table 2.** Mean average precision (mAP) of object detection task on UA-DETRAC (car)

| Inference dataset | Max accuracy of trained model | Baseline: Fine-tuning | Baseline2: Average | Gating Network: All (30 models) | Gating Network: Top 5 (5 models) |
|---|---|---|---|---|---|
| T1 | 85.6 (S1) | 72.6 | 87.9 | **90.4** | <u>**91.0**</u> |
| T2 | 78.6 (S1) | 78.6 | 67.6 | **80.5** | <u>**81.6**</u> |
| T3 | 90.7 (S25) | 85.8 | 87.0 | <u>**94.1**</u> | 93.7 |
| T4 | 81.3 (S2) | 80.5 | <u>**82.1**</u> | 81.0 | 82.0 |

**Baseline2 average:** According to the T2 dataset, the accuracy is the lowest of all the evaluation methods in Table 2. The accuracy of individual trained models toward T2 dataset is also overall low. In particular, the accuracies of model S5, S6, S14, and S24 are about under 20% (in Appendix B). Therefore, we supposes that, taking the average of all models, these low accuracy models could reduce the total accuracy. On the other hand, in the T4 dataset, the accuracy is the highest of all the methods, but there is no difference in accuracy. The cause of this result seems to be that training data T4' is not suitable for inference data. Training data T4' has few cars in the image, but inference data T4 has a lot of cars in the image. These differences of numbers of recognition objects are generally considered to be difficult test patterns for all methods.

**Gating network (All):** The method using the gating network achieves high accuracy. Gathering the knowledge of all the trained models seems to be effective for improving accuracy. It is considered that the optimal weight that is applied to all trained models is estimated for each inference data. Thus, even if there is few training data, we can increase the accuracy to utilize the existing trained models. Although not stated in Table 2, the accuracy for all training data (T1', T2', T3', and T4') is 92.75% on average. As mentioned in the result of fine-tuning section, in the case of fine-tuning, the accuracy for training data is 100% on average. This result is not considered to be over-fitting unlike with the fine-tuning results. Therefore, we assume that the gating network has a regularization effect.

**Gating network (top 5):** Once we train the gating network using all trained models, finally, we pick the top 5 trained models that have a large weight value manually (Table 3). That is to say, the trained model that is the most meaningful is selected for training data. As the training and inference data are similar, selected trained models are also suitable for inference data. Compared to the gating network (all) method, even the number of trained models has been reduced by 1/6, and the accuracy is almost the same or slightly improved. This result shows the accuracy remains good while suppressing the increase in computer resources.

**Table. 3.** Top 5 selected trained models and their weight value of the gating network.

| T1 | | T2 | | T3 | | T4 | |
| --- | --- | --- | --- | --- | --- | --- | --- |
| Selected model | Weight value | Selected model | Weight value | Selected model | Weight value | Selected model | Weight value |
| S2 | 41.2 | S1 | 44.6 | S2 | 41.2 | S1 | 44.6 |
| S1 | 14.1 | S2 | 14.1 | S1 | 14.1 | S2 | 14.1 |
| S11 | 10.9 | S15 | 8.1 | S11 | 10.9 | S15 | 8.1 |
| S15 | 5.5 | S10 | 6.6 | S15 | 5.5 | S10 | 6.6 |
| S29 | 4.5 | S19 | 4.0 | S29 | 4.5 | S19 | 4.0 |

Table. 4. The accuracy (mAP) when increasing the number of models to be fused

| Inference dataset | S1-S5 (5 models) | S1-S10 (10 models) | S1-S15 (15 models) | S1-S20 (20 models) | S1-S30 (30 models) |
|---|---|---|---|---|---|
| T1 | 87.9 | 87.1 | 89.0 | **90.7** | 90.4 |
| T2 | 79.2 | **81.7** | 81.1 | 81.3 | 80.5 |
| T3 | 66.5 | 70.0 | 81.7 | 91.2 | **94.1** |
| T4 | 80.9 | 80.5 | 80.2 | **81.3** | 81.0 |

**Effect of number of models on accuracy:** In the previous discussions, we chose the top 5 trained models according to the weight value. How does the accuracy change as the number of models is increased incrementally? In answering this question, we performed additional experiments. In Table 4, we show the results of adding the trained models. As an overall trend, adding the trained models tends to increase the accuracy. Focusing on the T3 dataset, the accuracy increases dramatically from 15 to 20 models. That is to say, the added models (S16, S17, S18, S19, and S20) contribute to the accuracy improvement. In particular, models S16 and S17 have accuracy within the top 3 of all the trained models in T3 (Appendix B). On the other hand, in the case of T2, adding the rest of the 10 models (from S21 to S30) has an adverse effect. The accuracies of the added models (from S21 to S30) are not good (Appendix B). Thus, adding a model with low accuracy will not contribute to improving the accuracy, but rather will lead to a bad result.

## 5    Conclusion

Existing transfer learning has focused on how to close the gap between a source domain and a target domain one-to-one. In this work, we have introduced and tackled a new scenario of transfer learning with various source domains. This scenario often arises in real applications, especially in video surveillance systems. In this situation, we propose a method to fuse many trained models using a gating network and apply these models to the object detection tasks. Our method is evaluated on the UA-DETRAC dataset and performs significantly better than the baseline methods. Safe and precise video surveillance system integration is one of the keys to social implementation. We have shown in this paper that it is possible. For future work, to further suppress the increase in the number of models, we will work on knowledge distillation technology.

# Appendices

## A  UA-DETRAC dataset

| Model | File name | Model | File name | Model | File name |
|---|---|---|---|---|---|
| S1  | mvi_20011 | S11 | mvi_40141 | S21 | mvi_40752 |
|     | mvi_20012 |     |           |     |           |
| S2  | mvi_20051 | S12 | mvi_40152 | S22 | mvi_40871 |
|     | mvi_20052 |     |           |     |           |
| S3  | mvi_39771 | S13 | mvi_40161 | S23 | mvi_41073 |
|     |           |     | mvi_40162 |     |           |
| S4  | mvi_39781 | S14 | mvi_40181 | S24 | mvi_63525 |
| S5  | mvi_39801 | S15 | mvi_40191 | S25 | mvi_63561 |
|     |           |     | mvi_40192 |     | mvi_63562 |
|     |           |     |           |     | mvi_63563 |
| S6  | mvi_39811 | S16 | mvi_40201 | S26 | mvi_39031 |
| S7  | mvi_39821 | S17 | mvi_40204 | S27 | mvi_39211 |
| S8  | mvi_39851 | S18 | mvi_40771 | S28 | mvi_39371 |
|     |           |     | mvi_40772 |     |           |
|     |           |     | mvi_40774 |     |           |
|     |           |     | mvi_40775 |     |           |
| S9  | mvi_39931 | S19 | mvi_40732 | S29 | mvi_39401 |
| S10 | mvi_40131 | S20 | mvi_40751 | S30 | mvi_39501 |

| Model | File name | Model | File name |
|---|---|---|---|
| T1 | mvi_20034 | T1' | mvi_20032 |
|    | mvi_20035 |     |           |
| T2 | mvi_20063 | T2' | mvi_20061 |
|    | mvi_20064 |     |           |
|    | mvi_20065 |     |           |
| T3 | mvi_63554 | T3' | mvi_63352 |
| T4 | mvi_40714 | T4' | mvi_40711 |

## B  Accuracy of trained model on test dataset

| Model | T1 | T2 | T3 | T4 | Model | T1 | T2 | T3 | T4 |
|---|---|---|---|---|---|---|---|---|---|
| S1 | 85.6 | 78.6 | 40.3 | 71 | S16 | 67.7 | 48.6 | 82.1 | 67 |
| S2 | 80.6 | 68.1 | 32.8 | 81.3 | S17 | 75.7 | 51.7 | 85.1 | 70.5 |
| S3 | 51.3 | 28.2 | 18.9 | 49.8 | S18 | 61.3 | 52.9 | 41.7 | 68.3 |
| S4 | 52.5 | 36.1 | 50.8 | 60.7 | S19 | 61.5 | 47.4 | 64.1 | 56.8 |
| S5 | 35.4 | 18.1 | 52.2 | 24.7 | S20 | 63.4 | 47.1 | 53 | 59.4 |
| S6 | 20.3 | 10.1 | 13.7 | 11.6 | S21 | 47.1 | 32.9 | 61.5 | 49.1 |
| S7 | 54.3 | 36.3 | 61.4 | 40.6 | S22 | 62.4 | 54.9 | 55.1 | 71.4 |
| S8 | 31.7 | 13.9 | 9.3 | 21.5 | S23 | 57 | 34 | 81 | 51 |
| S9 | 54.2 | 29.7 | 40.2 | 32 | S24 | 34.8 | 15.1 | 27.8 | 40.4 |
| S10 | 58.7 | 51.1 | 34.4 | 74.1 | S25 | 48.5 | 27.8 | 90.7 | 48.8 |
| S11 | 72.1 | 52 | 44.3 | 75.2 | S26 | 67.2 | 47.9 | 48.6 | 62.7 |
| S12 | 64.1 | 41.2 | 59.6 | 61.3 | S27 | 61.4 | 48.6 | 14.9 | 52.4 |
| S13 | 72.8 | 59.9 | 72.1 | 76.1 | S28 | 59.2 | 40.1 | 57.1 | 69.1 |
| S14 | 25.1 | 17.6 | 47 | 47.1 | S29 | 65.9 | 49.3 | 44.8 | 73.7 |
| S15 | 73.6 | 58.7 | 66.8 | 74.7 | S30 | 69.8 | 51.5 | 32.8 | 73 |